\documentclass[sigconf]{acmart}
\setcopyright{acmcopyright} 
\AtBeginDocument{%
  \providecommand\BibTeX{{%
    \normalfont B\kern-0.5em{\scshape i\kern-0.25em b}\kern-0.8em\TeX}}}

\copyrightyear{2020}
\acmYear{2020}
\setcopyright{iw3c2w3g}
\acmConference[WWW '21]{Proceedings of the Web Conference 2021}{April 19--23, 2021}{Ljubljana, Slovenia}
\acmBooktitle{Proceedings of the Web Conference 2021 (WWW '21), April 19--23, 2021, Ljubljana, Slovenia}
\acmPrice{}
\acmDOI{10.1145/3442381.3449892}
\acmISBN{978-1-4503-8312-7/21/04}
\begin{document}
\title{Quiz-Style Question Generation for News Stories}

\author{Adam D. Lelkes}
\authornote{Both authors contributed equally to this research.}
\affiliation{\institution{Google Research}}
\email{lelkes@google.com}

\author{Vinh Q. Tran}
\authornotemark[1]
\affiliation{\institution{Google Research}}
\email{vqtran@google.com}

\author{Cong Yu}
\affiliation{\institution{Google Research}}
\email{congyu@google.com}

\begin{abstract}
A large majority of American adults get at least some of their news from the Internet.
Even though many online news products have the goal of informing their users about the news,
they lack scalable and reliable tools for measuring how well they are achieving this goal,
and therefore have to resort to noisy proxy metrics (e.g., click-through rates or reading time) to track their performance.

As a first step towards measuring news informedness at a scale, we study the problem of quiz-style multiple-choice question generation, which may be used to survey users about their knowledge of recent news. In particular, we formulate the problem as two sequence-to-sequence tasks: question-answer generation (QAG) and distractor, or incorrect answer, generation (DG).
We introduce \emph{NewsQuizQA}, the first dataset intended for quiz-style question-answer generation, containing 20K human written question-answer pairs from 5K news article summaries. Using this dataset, we propose a series of novel techniques for applying large pre-trained Transformer encoder-decoder models, namely PEGASUS and T5, to the tasks of question-answer generation and distractor generation.

We show that our models outperform strong baselines using both automated metrics and human raters.
We provide a case study of running weekly quizzes on real-world users via the Google Surveys platform over the course of two months. We found that users generally found the automatically generated questions to be educational and enjoyable.
Finally, to serve the research community, we are releasing the \emph{NewsQuizQA} dataset.

\end{abstract}


\maketitle

\section{Introduction}
Americans are increasingly relying on online news products, such as news publisher websites, news aggregators, and social media to get their news~\cite{geiger_2020}.
Efforts to understand and improve the health of the news ecosystem for these consumers can be thought of as contributing to one of three directions:  user satisfication, media literacy, and news informedness. 
User satisfication measures how happy readers are with the content or product that they interact with, while media literacy measures whether the readers have the ability to correctly assess the news they are consuming. News informedness measures instead how much readers know and understand the important current events that are in the news. This is the least studied of the three and although many online news products have the goal of informing their users about the news, they generally have lacked scalable and reliable tools for measuring how well they are achieving this goal---often relying on engagement metrics as a proxy for informedness. Recently, academic works in news informedness have used multiple-choice and true-false questions to assess respondents' current events, civics, and general background knowledge~\cite{pew_2020, aba_2020, cfr_2019}. However, these questions are manually written, resulting surveys that target only evergreen knowledge or that run very infrequently.

\begin{figure}[t!]
  \centering
  \includegraphics[width=\linewidth]{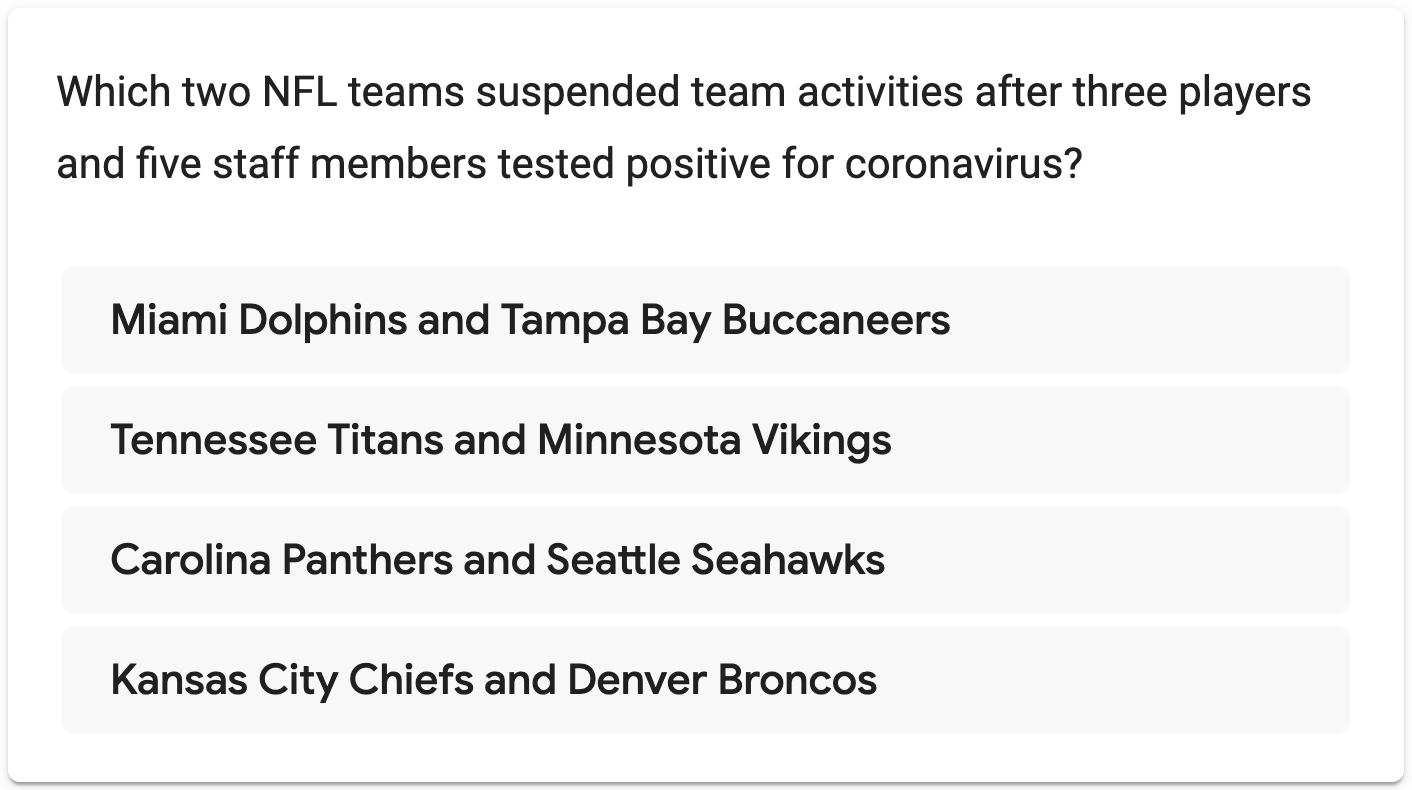}
  \caption{Example of a quiz-style multiple-choice question automatically generated by our sequence-to-sequence models. Question and distractors were selected by a human curator on September 30, 2020 as described in Section~\ref{case_study}, with no edits made to the original generated text. (The correct answer is "Tennessee Titans and Minnesota Vikings.")}
  \Description{Question: Which two NFL teams suspended team activities after three players and five staff members tested positive for coronavirus? Answer option 1: Miami Dolphins and Tampa Bay Buccaneers. Answer option 2: Tennessee Titans and Minnesota Vikings. Answer option 3: Carolina Panthers and Seattle Seahawks. Answer option 4: Kansas City Chiefs and Denver Broncos.}
  \label{fig:example}
\end{figure}

As a first step towards measuring news informedness at scale, we study the problem of quiz-style multiple-choice question generation, which may be used to survey users about their knowledge of recent news. We formulate multiple-choice question generation as two sequence-to-sequence tasks: \emph{quiz-style} question-answer generation and \emph{distractor} (incorrect answer option) generation. For quiz-style question-answer generation, given an input passage, the goal is to generate a question and its associated correct answer (also known as the \emph{key}), with the requirement the question must be able to stand independently, e.g. in a pop quiz, without reference to any source material. In other words, the question must contain enough information to be answered without association to any passage. For distractor generation, given the input passage, question, and correct answer, we aim to generate incorrect answer options that are both plausible answers to the question, and clearly distinct from the correct answer and from each other.

There has been significant progress recently in question generation, most of which have been based on the sequence-to-sequence attention framework of Du et al.~\cite{du-etal-2017-learning} However, many of these works have either assumed that (1) the answer is not needed or is provided as input, or (2) the question is asked in reference to a readily available source material, whether it be text, image, dialogue, or otherwise~\cite{rajpurkar-etal-2016-squad, mostafazadeh-etal-2016-generating, reddy-etal-2019-coqa}. Accordingly, many existing works have used reading comprehension datasets such as SQuAD~\cite{rajpurkar-etal-2016-squad} for question generation. This approach does not work for quiz-style question generation, as reading comprehension questions assume that the reader has access to the source passage. Previous works that have used questions derived from search engine queries, such as Natural Questions or MS Marco, are also unsuitable for this task as question-like queries are often too short, and not fluent enough for the quiz context~\cite{47761, DBLP:journals/corr/NguyenRSGTMD16}.

To address this, we devise a new dataset, \emph{NewsQuizQA}, intended for quiz-style question-answer generation. \emph{NewsQuizQA} consists of 20K human written question-answer pairs derived from 5K news article summaries, covering news from June 2018 to June 2020. Writers for this task were instructed specifically to create questions that could stand alone, without reference to the source text. Each news article summary has four associated reference question-answer pairs, all of which are considered to be correct possible outputs.

Using this dataset, we propose a series of techniques for applying large pre-trained Transformer encoder-decoder models, PEGASUS and T5, for question-answer generation and distractor generation, respectively. In order to efficiently learn question-answer generation from the relatively small \emph{NewsQuizQA} dataset, we leverage existing question-answer datasets via weak supervision in a \emph{mid-training} stage, as well as apply a min-loss function, which we call \emph{minimum reference loss}, to better leverage the multiple, diverse, reference question-answer pairs associated with each passage.

For distractor generation, we propose a novel formulation of distractor generation as sampling from a closed-book question-answering model. We show that we are able to generate an abundance of high quality question-answering data by applying our own question-answer generation model to a large corpus of documents, and that we can use this data to fine-tune T5 to generate better distractors for our questions. We also propose a simple heuristic to mitigate duplicates between our distractor generator samples and the correct answer.

We evaluate our methods using both automatic metrics and human evaluation and show that they outperform strong baselines. In doing so we identify issues with naively using ROUGE to evaluate question-answer generation, and propose an extension of ROUGE that is more suitable. Finally, we used our models to automatically generate weekly surveys, which were run on real users via the Google Surveys platform over the course of two months. The results of this case study suggest that our generated multiple choice questions are both educational and enjoyable. 
In summary, this paper provides a diverse list of technical contributions:
\begin{enumerate}
  \item We propose studying task of multiple-choice quiz question generation for news articles, and construct the first quiz-style question-answer generation dataset, \emph{NewsQuizQA}.
  \item For question-answer generation, to improve performance in the low-resource QAG setting, we propose a weakly supervised mid-training phase that leverages existing QA datasets.
  \item We demonstrate the effectiveness of a min-loss function for learning sequence-to-sequence tasks where for every input there are many correct, diverse, reference outputs.
  \item We propose an approach to distractor generation via closed-book question-answering, and adapt an existing QA system to our news quiz domain by training on a large corpus of automatically generated question-answer pairs.
  \item We provide an extension of ROUGE useful for evaluating pairs of generated text, i.e. question-answer pairs.
\end{enumerate}

\section{Related Work}\label{relatedwork}
In this section, we briefly summarize recent work on news informedness and automated generation of multiple-choice quiz questions. For a comprehensive review of the field, we refer the reader to~\cite{kurdi2020systematic} as well as earlier surveys such as~\cite{ch2018automatic} and~\cite{papasalouros2018semantic}.

\textbf{News informedness} has been studied in recent years, primarily using surveys consisting of manually written multiple-choice or true-false questions about current events~\cite{pew_2020, jenkins_2018}, civic knowledge~\cite{aba_2020, annenberg_2020, acta_2019}, or general background knowledge~\cite{cfr_2019}. There has also been work showing the advantages of the multiple-choice question format for assessing knowledge~\cite{10.2307/2669369}, and the disadvantage of self-assessing knowledge due to the Dunning-Kruger effect~\cite{doi:10.1111/pops.12490}.

\textbf{Multiple choice question (MCQ) generation} has been studied since the late '90s, however existing MCQ systems have primarily relied on heuristics and rule-based transformations~\cite{8585151}. Previous works have also studied MCQ question generation as transformations from structured data such as knowledge graphs~\cite{DBLP:journals/corr/SeylerYB16, Alsubait2014}. To the best of our knowledge, there have been no previous works or systems to frame the entirety of MCQ generation as a sequence-to-sequence learning task.

\textbf{Question generation} however has gained interest and has made much progress in recent years. While very related to question-answer generation, question generation aims to only generate a question given an input and optionally a pre-selected answer (answer-focused question generation). While early work utilized rule-based transformations and an over-generate and rank approach~\cite{heilman-smith-2010-good}, recent mainstream works rely on neural-network based sequence-to-sequence methods, in particular encoder-decoder models with attention~\cite{du-etal-2017-learning, zhao-etal-2018-paragraph}.
There have also been various extensions, including copy or pointer mechanisms to overcome out-of-vocabulary issues~\cite{du-cardie-2018-harvesting, sun-etal-2018-answer}, mechanisms for handling paragraph-length inputs~\cite{zhao-etal-2018-paragraph, zhang-bansal-2019-addressing}, and more~\cite{nema-etal-2019-lets, jia-etal-2020-ask}. Of particular interest to this work is the recent success of large-scale pre-training for Transformer-based~\cite{NIPS2017_7181} encoder-decoder models on many NLP tasks, such as BERT~\cite{devlin2019bert}, T5~\cite{JMLR:v21:20-074}, and PEGASUS~\cite{zhang2020pegasus}. UniLM~\cite{DBLP:journals/corr/abs-1905-03197} and UniLMv2~\cite{bao2020unilmv2} have reported the highest scores for question generation via SQuAD, although many state-of-the-art pre-trained language models do not report their scores on this task. Additionally, leveraging weak supervision via an intermediate training stage to further refine pre-trained models have also been explored in previous works in question generation and headline generation~\cite{narayan2020qurious, 49098}.

{\small
\begin{table*}[t]
  \caption{Comparison with other selected QA datasets. \emph{NewsQuizQA} has the longest question and answers, in  characters.}
  \begin{tabular}{lcccccc}
    \toprule
    \textbf{Dataset} & \textbf{Domain} & \textbf{Type} & \textbf{QA Pairs} & \textbf{Avg. pairs per passage} & \textbf{Avg. question length} & \textbf{Avg. answer length}\\
    \midrule
    SQuAD & Wikipedia & Reading comprehension & 98,169 & 4.68 & 59.6 & 20.1 \\
    Natural Questions & Wikipedia & Search query & 99,486 & 1.55 & 48.3 & 24.9 \\
    NewsQA & News & Reading comprehension & 90,529 & 7.50 & 37.0 & 25.5 \\
    NewsQuizQA & News & Pop quiz & 20,024 & 4.00 & 77.8 & 27.1 \\
    \bottomrule
  \end{tabular}
  \label{tab:dataset_comparison}
\end{table*}
}

\textbf{Distractor generation} papers have primarily focused on ranking a set of distractor candidates, often focusing on single words or entities as distractors. Ranking was often done based on similarity to the key, for instance using handcrafted similarity measures~\cite{10.5555/2043132.2043139} or word embeddings~\cite{kumar2015revup}~\cite{10.5555/3061053.3061140}, or based on learned ranking models~\cite{liang-etal-2018-distractor}. There has also been work that used GANs for distractor generation~\cite{liang2017distractor}.

\textbf{Datasets} exist for question generation; however, none are suitable for our quiz-style question-answer generation task. SQuAD is the most commonly used for training and evaluation for answer-focused question generation~\cite{rajpurkar-etal-2016-squad}. However since it is a reading comprehension dataset, many questions contain references to the source text that makes it unsuitable for generating quiz-style questions. NewsQA~\cite{trischler-etal-2017-newsqa} is another reading comprehension dataset and while it is in the news domain, it faces similar issues to SQuAD in terms of referencing source text. QA datasets derived from search-engine queries such as Natural Questions~\cite{47761} and MS Marco~\cite{DBLP:journals/corr/NguyenRSGTMD16} do contain questions that could stand alone in a quiz, however these questions often resemble search queries: lacking capitalization, short and non-descriptive, and often leaving out articles or other words that limit fluency. TriviaQA~\cite{joshi-etal-2017-triviaqa} is a dataset that provides high-quality trivia questions. However, noisy associations to source text limits its viability for the purpose of question-answer generation. Other datasets for question generation not listed here either completely lack source passage associations or are designed for other tasks such as clarification or conversational QA~\cite{10.1145/2661829.2661908, reddy-etal-2019-coqa, kumar-black-2020-clarq}.

\section{Problem Description}
Given an input passage $I$, i.e. a news article or a summary of a news article, we aim to generate a quiz-style multiple choice question which consists of a question $Q$, an answer $A$, and a set of distractors, incorrect answer options, $D_1, D_2, \hdots , D_k$ (in most settings $k=3$). For the question to be considered quiz-style, $Q$ has the additional requirement that it must be able to be answered without assuming that the reader has access to any other source material. For example, certain reading comprehension questions containing direct references such as "according to the passage" or dangling mentions such as "what did \emph{she} say?" would not be considered quiz-style. To encourage natural and interesting questions, other than that it must be the correct answer for $Q$, there are no additional constraints on $A$, which can be any span of text and can even be abstractively generated from $I$. The distractors $D_1, \hdots , D_k$ have the requirements that they must be (1) incorrect, (2) plausible answers for $Q$, and (3) unique from each other and also from $A$. In general, we aim to generate questions that not only meet these requirements, but are also educational and enjoyable to complete. 

For the rest of the paper we will approach this multiple-choice question generation problem as two separate sequence-to-sequence tasks: question-answer generation (QAG) and distractor generation (DG). QAG expects $I$ as input and generates the question-answer pair $(Q, A)$, while DG expects $I$, $A$, and $Q$ as input as generates distractors $D_1, \hdots, D_k$.

\section{\emph{N\lowercase{ews}Q\lowercase{uiz}QA} Dataset}

To address the lack of viable training data for the task of quiz-style QAG, we introduce a new dataset called \emph{NewsQuizQA}. The dataset contains 20K human written quiz-style question and answer pairs coming from 5k news articles between June 2018 to June 2020. A proprietary clustering algorithm iteratively loads articles published in a recent time window and groups them based on content similarity. To construct the dataset, on a weekly basis the top 50 clusters are taken and for each cluster a representative article close to the centroid is selected. For each article a summary is generated using a PEGASUS model fine-tuned on the CNN/Dailymail summarization dataset ~\cite{10.5555/2969239.2969428}, a state-of-the-art model for single document summarization. Summaries are used for more efficient data collection over using entire news articles, which can be long and time consuming to read for human writers. Each summary is then given to five human writers who are asked to read the passage and write a question and answer pair that abides the following rules:

\begin{enumerate}
  \item The question is answerable based on information from the passage only.
  \item The question can stand alone without the passage. I.e., it provides enough background to understand what is being asked, without the passage.
  \item The question has a short answer. I.e., not “how” or “why” type questions that can only be answered by a full sentence.
  \item The question should be about one of the most interesting or important aspects of the passage.
  \item The question ends with a question mark.
  \item The question is not a "Yes" / "No" question.
  \item The answer is a word or short phrase with no ending punctuation.
  \item The answer conveys only the answer, does not contain unnecessary parts, and does not restate parts of the question.
\end{enumerate}

\begin{figure}[t!]
  \centering
  \minipage{0.5\linewidth}
    \centering
    \includegraphics[width=\linewidth]{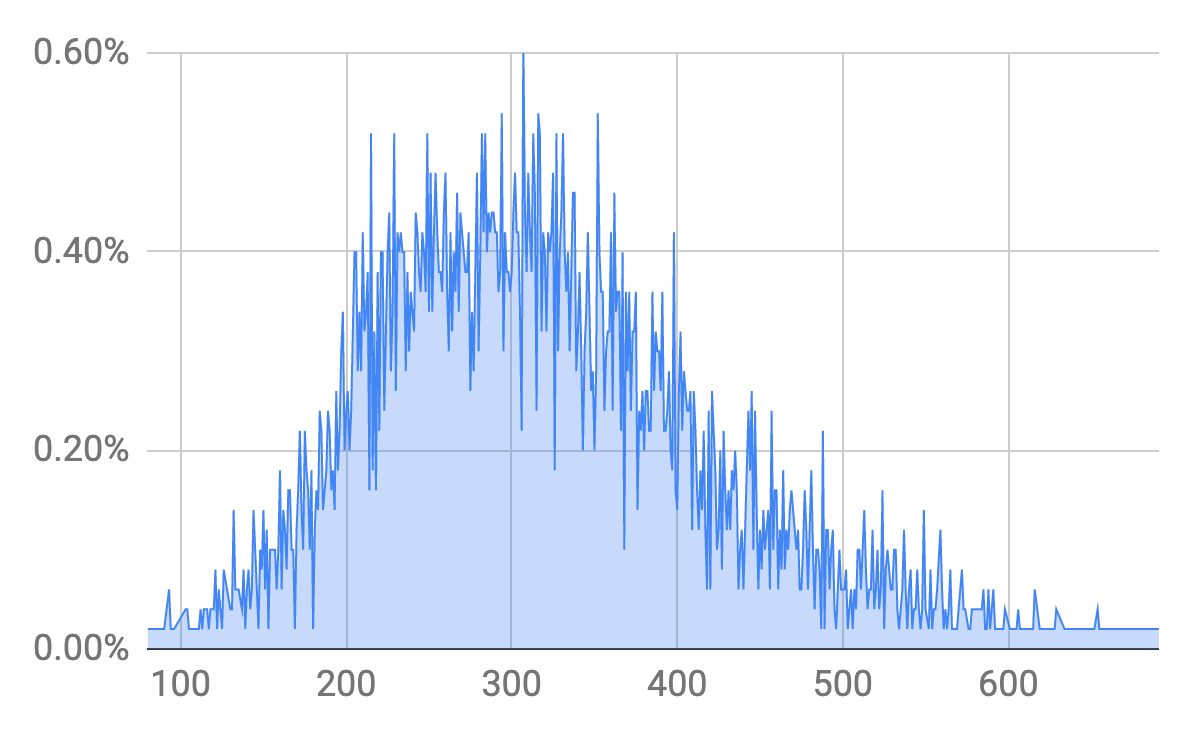}
    \small{(a) Summary Length (chars.)}
  \endminipage\hfill
  \minipage{0.5\linewidth}
    \centering
    \includegraphics[width=\linewidth]{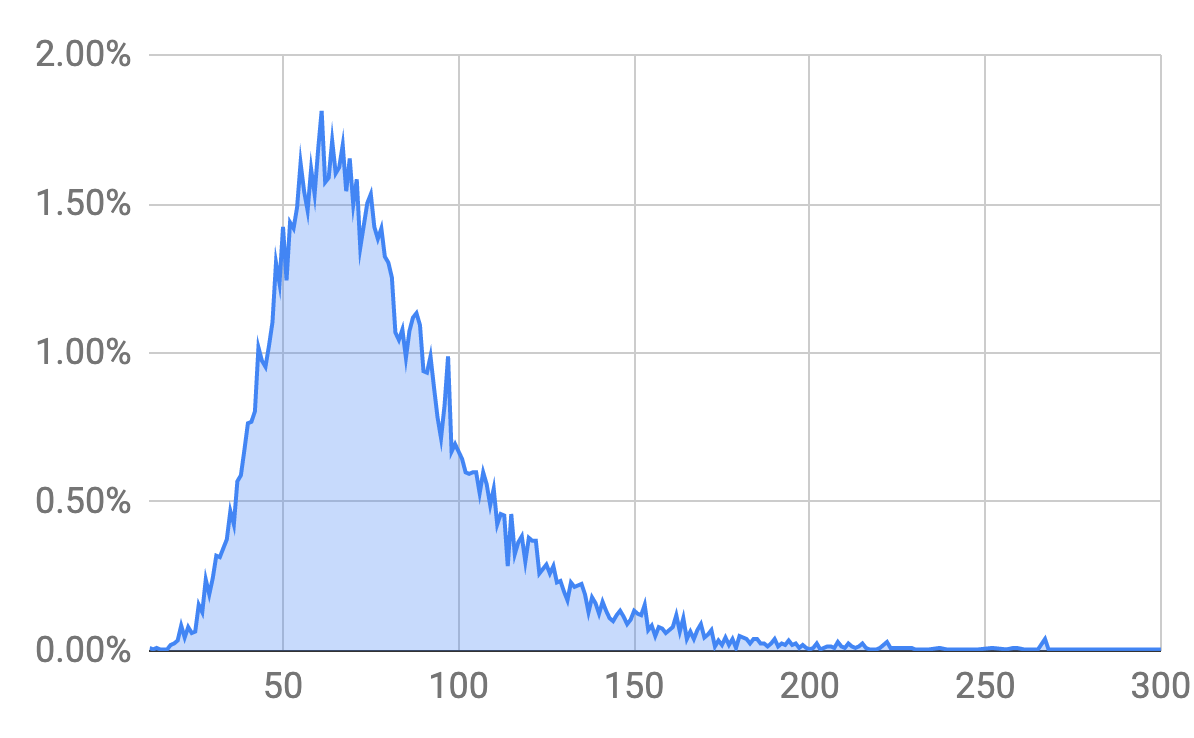}
    \small{(b) Question Length (chars.)}
  \endminipage\hfill
  \minipage{0.5\linewidth}
    \centering
    \includegraphics[width=\linewidth]{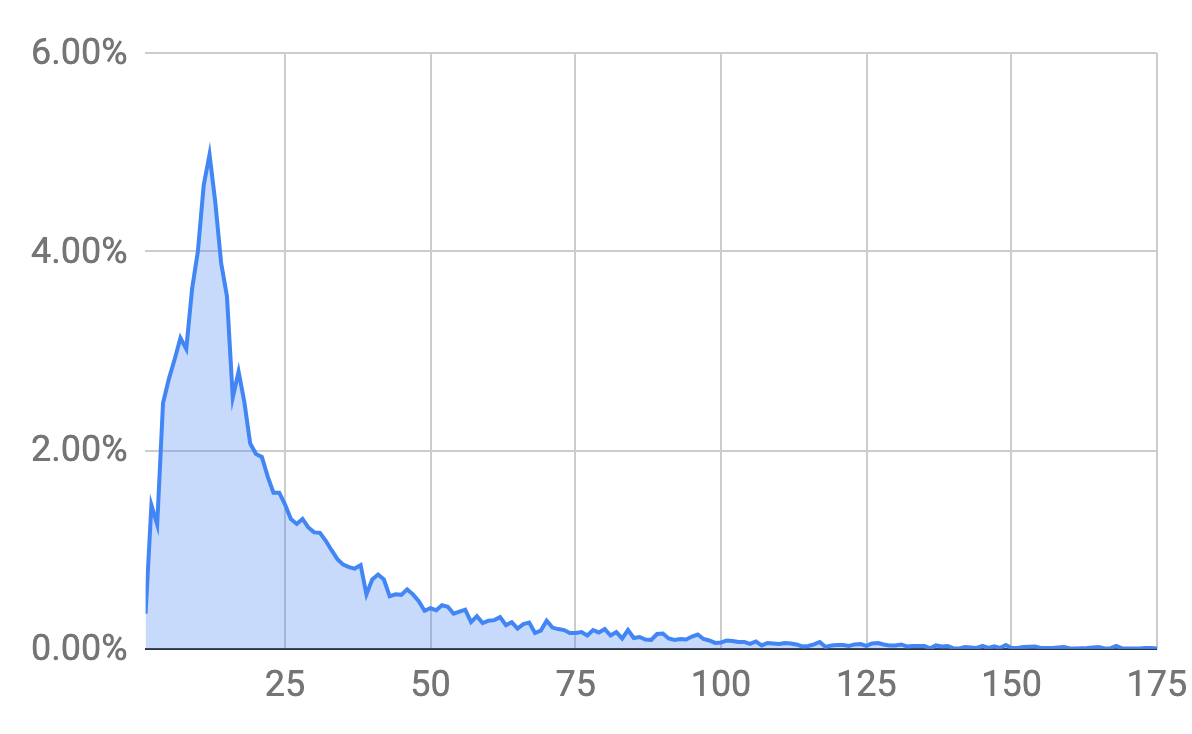}
    \small{(c) Answer Length (chars.)}
    \endminipage\hfill
  \minipage{0.5\linewidth}
    \centering
    \includegraphics[width=\linewidth]{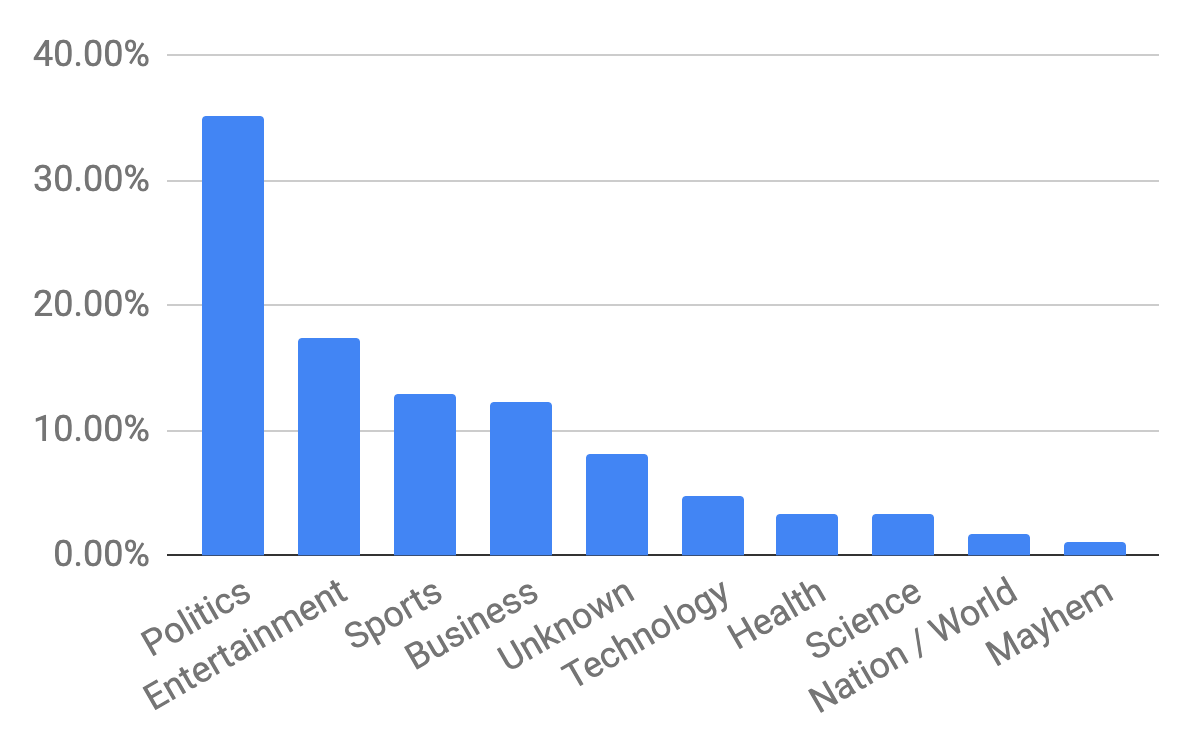}
    \small{(d) Topic Distribution of Articles}
  \endminipage\hfill
  \caption{Visualization of \emph{NewsQuizQA} data statistics.}
\end{figure}

From this process, we collect exactly 26,000 question-answer pairs from 5,200 news article summaries. We then apply a post processing step to mitigate low quality questions. Namely, the human-written questions are processed via a state-of-the-art grammar error correction model~\cite{lichtarge-etal-2019-corpora} in order to fix minor grammar, spelling, punctuation, and capitalization errors. Then, the shortest question written for each summary is removed, as well as any question containing specific blocklisted phrases such as "I" or "According to the passage." Aftwards, only the summaries with exactly four human written question-answer pairs are kept for the final dataset. An 80-10-10 split is then randomly sampled from the 5k summaries to produce training, validation, and test sets, respectively.

The result of this process is a high quality dataset containing questions that are able to be presented by themselves, without an associated passage in a quiz setting. We observe that these questions are often longer (Table~\ref{tab:dataset_comparison}), and capture more information from the source passage in their writing. The questions also span various forms and answer types beyond simply "who" or "where" questions.

Two practical challenges arise due to this dataset. First, since the cost of question creation is expensive, the size of the dataset is quite small in the number of news article summaries covered. Second, while there are few summaries there are many diverse and equally correct reference question-answer pairs for each summary. How do we best use these multiple diverse reference outputs during training? In the next section we will present the techniques we use to address both of these challenges.

A majority of the NewsQuizQA dataset can currently be accessed \href{https://github.com/google-research-datasets/NewsQuizQA}{here}\footnote{\href{https://github.com/google-research-datasets/NewsQuizQA}{https://github.com/google-research-datasets/NewsQuizQA}}. A remaining portion of NewsQuizQA is currently under review and additional question-answer pairs will be released in subsequent versions of the dataset.

\section{Question-Answer Generation}\label{qag}

In this section we will elaborate on the techniques we use to learn a high performing QAG model for the \emph{NewsQuizQA} dataset.

\subsection{Model Architecture}

In order to leverage recent improvements in natural language generation brought on by large pre-trained Transformer encoder-decoder models, we base our QAG model on PEGASUS, which recently achieved state-of-the-art performance on single document summarization. We begin with a large 568M parameter "Mixed and Stochastic" PEGASUS model, which was trained for 1.5M steps on both C4 \cite{JMLR:v21:20-074} and HugeNews \cite{zhang2020pegasus} using the Gap Sentences Generation pre-training objective. We hypothesized that PEGASUS would be well suited for paragraph-level question-answer generation since QAG shares many similarities to abstractive summarization. However, we will see in Section~\ref{results}, that high performance on this task may not be limited to only PEGASUS.

\subsection{Mid-training via Weak Supervision}\label{mid-train}
Although PEGASUS has demonstrated strong fine-tuning performance in low-resource settings for the summarization task, we aim to further adapt PEGASUS to our low-resource QAG setting. We propose training PEGASUS first on weakly supervised data from existing QA datasets. We denote this as the \emph{mid-training} phase. In particular, we combine SQuAD, Natural Questions, and NewsQA and first fine-tune PEGASUS on this combined dataset. Although individually none of these datasets are suitable for our application, the combined examples do cover several qualities that are desired: length, quality, domain, and self-containment.

One challenge with naively combining these datasets is that two very different target QA pairs from two different datasets may come from very similar input passages. This is especially true for SQuAD and Natural Questions as both are derived from Wikipedia, but have very different question writing styles: long well-formed reading comprehension questions versus short uncased questions from search queries. To address this issue, we borrow inspiration from T5 and prefix all inputs with a label denoting the style we expect the model to generate. e.g. \texttt{"Style SQuAD:"}, \texttt{"Style NQ:"}, \texttt{"Style NewsQA:"}. The model then is able to learn to associate certain styles of generated questions with certain labels in the input and does not have to infer this information from the contents of the input, which may be prone to overfitting. These datasets also contain multiple correct output reference per inputs, which we disaggregate and shuffle during this mid-training step.

\subsection{Minimum Reference Loss}\label{minrefloss}
The \emph{NewsQuizQA} dataset contains four reference questions for each input passage.
These reference questions are typically diverse, with questions often asking about different aspects of the story with different answers.
We evaluate our models based on how well the predicted question-answer pairs match \emph{any} of the reference outputs; i.e., for each input, we would like to optimize for the maximum score across the reference outputs. (See Section~\ref{evalmetrics} for more details on evaluation metrics.)

Natural language generation settings with multiple correct reference outputs have been studied before; e.g., in the machine translation literature where often there are multiple correct translations of the same input. Much of that work has focused on model evaluation and data augmentation. One recent work~\cite{zheng-etal-2018-multi} compared various training strategies given multiple references; each of the methods involved disaggregating the reference outputs and converting the dataset to a single reference dataset in different ways (such as by sampling from the reference outputs or shuffling them).

Inspired by a recent work ~\cite{jia-etal-2020-ask}, which applied a min-loss function to learn paraphrased reference outputs to encourage diversity in generation, we apply a min-loss function instead directly to the golden reference outputs provided by the dataset. Specifically, we propose the following method for training with diverse reference outputs.
During training, for each input passage and in each training step, we compute the losses for the model's predictions for all reference question-answer outputs.
Since training is done by teacher forcing, this means there is a cross-entropy loss for the next token prediction for each prefix of each of the $R$ reference outputs, for a total of $RT$ losses, where $T$ is the maximum output length. Let $L\in\mathbb{R}^{R \times T}$ denote this loss matrix and let $l_i$ denote the length of the $i$th prediction in tokens. We then define the loss for the example as

$$L= \min_{i\in{1 \ldots R}} \frac{1}{l_i} \sum_{j=1}^T L_{i,j}$$

\noindent This is equivalent to choosing a single reference output for each input, but instead of choosing it at the beginning of training, we choose it dynamically at the end of each training step, always picking the reference output on which the model performs best, normalized for length.
To implement this, all references outputs for an example are always packed into the same batch. 

We show in Section~\ref{results} that this method empirically outperforms other multi-reference training strategies.

\section{Distractor Generation}
The primary goal of distractor generation is generating answer options that are plausibly answers to the question, and might appear correct to a user who does know the correct answer. Distractors should also be clearly distinct from the key and each other and they should not be correct answers to the question (for questions that might have multiple correct answers). The task of generating plausible distractors is made challenging by the fact that plausibility of a wrong answer option is inherently somewhat subjective, ill-defined, and dependent on the domain and context, as well as by the dearth of training and evaluation data. The most commonly used question answering datasets only contain the correct answer; available multiple-choice question datasets that include distractors, such as~\cite{lai2017race}~\cite{welbl2017crowdsourcing}, typically have short distractors and/or focus on reading comprehension or a narrow domain.

As we noted in Section~\ref{relatedwork}, most previous papers attempted to rank distractors based on various notions of similarity to the key, with the underlying assumption that answer option candidates that are syntactically and semantically close to the key could also pass for potentially correct answers. The downside of this approach is that similarity to the key does not always imply plausibility as an answer option and, even if it did, similarity as an answer option is very sensitive to the context of the question itself, in a way that is not likely to be captured by generic word or entity embeddings.

\subsection{Distractor Generation Model}
We take the novel approach of directly simulating what one would guess as the answer to the quiz question. To the best of our knowledge, we are the first ones to propose a model that can generate high quality distractors purely based on the question text itself, without depending on the key or the source passage. Moreover, unlike most previous work, our distraction generation model can generate free form text and, whether the question expects an entity, phrase, number, or even a whole sentence as the answer, the model is generally able to generate distractors that are syntactically and semantically appropriate (without knowing the correct answer's format).

Our approach was inspired by~\cite{roberts2020much} which demonstrated the ability of T5~\cite{JMLR:v21:20-074} to perform what they call "closed-book" question answering; i.e., fine-tuning a model that can answer questions without access to an external context or knowledge beyond the question itself. We observe that this setting closely resembles the setting of a human user attempting a (non-multiple choice) quiz question.

Similarly to~\cite{roberts2020much}, we fine-tuned T5 to predict the answer given just the question text itself. However, we used our own questions and answers for fine-tuning data. Given a question-answer generation model described in Section~\ref{qag}, we can generate arbitrary numbers of training examples for distractor generation without the need for human supervision, simply by running the QAG model on a large number of news article summaries. In practice, we are using T5-Large fine-tuned on little under 2M question-answer examples.

\subsection{Distractor Deduplication}\label{dgdedup}
To obtain distractors for a given question using the above model, we sample from the decoder of the model. Since we sample answer predictions indepedently, there is the risk of getting duplicate answer options.

We use the following heuristic to ensure answer option diversity. We first independently sample 4 times as many distractor candidates as the target number of distractors. Then we iteratively add a distractor from the set of candidates until we have the intended number of distractors. In each iteration, we select the distractor which is farthest from the set of already chosen answer options (including the key), according to a BERT-based text embedding model.

\section{Experiment Settings}

Question-answer generation and distractor generation are two very different tasks that require different evaluation approaches. In this section, we describe our evaluation methods for both tasks. For QAG, we rely on automatic metrics against the NewsQuizQA test set which we treat as golden data, while for DG we design a human evaluation task for measuring distractor quality. For both, we compare our methods against strong baselines. Afterwards, to further investigate the holistic quality of our generated multiple-choice questions, in Section~\ref{case_study} we evaluate our questions using surveying techniques on real users.

\subsection{Baseline Methods for QAG}

Although no previous methods are directly applicable to the problem of QAG, given the success of large pre-trained language models on the problem of question generation, we compare our model to the recent state of the art sequence-to-sequence model T5. We additionally provide an ablation study to evaluate the effect of our weakly supervised QAG mid-training and minimum reference loss. In total our baseline methods are as follows:

\begin{itemize}
    \item \textbf{T5 Disaggregated}. We fine-tune three versions of T5: Base, Large, and 3B. Each have approximately 220M, 770M, and 2.8B parameters respectively. To handle multiple correct target references, we follow previous work \cite{zheng-etal-2018-multi} in machine translation and disaggregate the reference targets, and then shuffle the single-reference dataset on every training epoch. 
    \item \textbf{PEGASUS Disaggregated}. We fine-tune PEGASUS, using the same disaggregation strategy as above. For reference, PEGASUS has 568M parameters.
    \item \textbf{PEGASUS Sample}. We fine-tune PEGASUS, however instead of disaggregate-and-shuffle, for each example, we sample one target reference a priori, resulting in a training set that is four times smaller. This baseline is intended to test whether using multiple target references in training as previous works did for machine translation may be detrimental in the QAG setting, where references are more diverse. 
    \item \textbf{PEGASUS + MinRefLoss}. We fine-tune PEGASUS using minimum reference loss as described in Section~\ref{minrefloss}. For comparison, we also include a variant without the length normalization factor in the minimum reference loss.
    \item \textbf{PEGASUS + Mid-training + MinRefLoss}. This is our complete QAG model. We first train PEGASUS on existing QA datasets using techniques described in Section~\ref{mid-train} for 100k steps, then fine-tune this model using minimum reference loss as described in Section~\ref{minrefloss}. We also include a variant without length normalization here. 
\end{itemize}

\noindent \textbf{Reproduction Details.} All models were fine-tuned for 20k steps. We found that training beyond 20k steps did not improve performance. Beam search with 8 beams and a beam alpha of 0.9 was used at prediction time when evaluating for all models. Checkpoints were selected based on validation set performance (R2-F-QAG described in next section) before evaluation against test set. PEGASUS and T5 models were trained with a learning rate of 0.00001 and 0.003 respectively. Batch sizes were picked to be as large as possible given memory constraints. For PEGASUS this was 128 for MinRefLoss, and 256 for Disaggregated and Sample. For T5 this was 128, 64, and 16 for Base, Large, and 3B respectively.
For question-answer generation, we used a maximum output token length of 128 for the combined question-answer output sequence.
For distractor generation, we used a maximum output token length of 64.
(There was no minimum output length.)
For all other hyperparameters, we used the default values.

\subsection{Evaluation Metrics for QAG}\label{evalmetrics}
We use ROUGE~\cite{lin2004rouge} scores for evaluating how well our generated text matches the reference outputs. ROUGE scores, the most commonly used metric for evaluating natural language generation models, measure n-gram overlap between the generation and the reference output.
In particular, we use ROUGE2-F1 (abbr. R2-F), the F1 scorer of bigram overlap between the generated text and the reference.
We note that when there are multiple reference outputs for an input example, as is the case for the \emph{NewsQuizQA} dataset, the ROUGE score of a prediction for an input example is defined as the maximum ROUGE score for the prediction across the reference outputs.

Question-answer generation differs from common natural language generation tasks (such as summarization or machine translation) in that the reference outputs always consist of two components, each of which are crucially important to the quality of the generation: the question and the answer.
While we model QAG as a combined sequence-to-sequence task, with a single output sequence containing both the question and the answer, we want to capture the structure of the task in our evaluation metrics.

For this, we adapt ROUGE scores to our use case in the following way: instead of aggregating n-gram overlap over both the question and the answer, for each reference output we compute the ROUGE score for the question and answer separately, and take the harmonic mean. We define this score as 0 if the prediction does not split into question and answer components. This version of ROUGE penalizes models that get either the question or the answer wrong more severely than classic ROUGE. We call this version of ROUGE scores ROUGE-QAG. We use ROUGE2-F1-QAG (abbr. R2-F-QAG) on the validation set for model selection. We additionally report the average length of the combined question and answer predictions.

{\small
\begin{table*}[ht!]
\caption{Performance comparison of different methods for QAG task on the \emph{NewsQuizQA} dataset test split.}
  \begin{tabular}{lccccccc}
    \toprule
    \textbf{Method} & \textbf{R1-F-QAG} & \textbf{R2-F-QAG} & \textbf{RL-F-QAG} & R1-F & R2-F & RL-F & \textbf{Avg. Length} \\
    \midrule
    T5 Disaggregated (Base) & 0.503 & 0.310 & 0.486 & 0.674 & 0.508 & 0.600 & 117 \\
    T5 Disaggregated (Large) & 0.512 & 0.325 & 0.490 & 0.693 & 0.530 & 0.618 & 117 \\
    T5 Disaggregated (3B) & 0.523 & 0.335 & 0.502 & 0.706 & 0.544 & 0.630 & 117 \\
    PEGASUS Disaggregated & 0.478 & 0.335 & 0.464 & 0.650 & 0.525 & 0.584 & 154 \\
    PEGASUS Sample & 0.506 & 0.331 & 0.487 & 0.696 & 0.541 & 0.622 & 108 \\
    PEGASUS + MinRefLoss (No Length Norm.) & 0.549 & 0.355 & 0.532 & 0.723 & 0.567 & 0.648 & 75.3\\
    PEGASUS + MinRefLoss & 0.544 & 0.360 & 0.528 & 0.729 & 0.580 & 0.653 & 75.0\\
    PEGASUS + Mid-training + MinRefLoss (No Length Norm.) & 0.556 & 0.363 & 0.539 & 0.728 & 0.576 & 0.653 & 61.8\\
   \textbf{PEGASUS + Mid-training + MinRefLoss} & \textbf{0.567} & \textbf{0.378} & \textbf{0.549} & \textbf{0.745} & \textbf{0.596} & \textbf{0.669} & \textbf{96.7} \\
    \bottomrule
    \label{tab:qag}
  \end{tabular}
\end{table*}
}

{\small
\begin{table}[ht!]
\caption{Human evaluation for distractor generation models.}
\label{tab:dgeval}
\begin{tabular}{lcc}
\toprule
\textbf{Statistic} & \textbf{Proposed method} & \textbf{Baseline} \\
\midrule
Plausible options per question & $1.5$ & $1.51$ \\
Single plausible option & $69.1\%$ & $69.3\%$ \\
Single plausible is a distractor & $46.4\%$ & $42.4\%$ \\
Keys not plausible & $39.2\%$ & $37.6\%$ \\
Distractors plausible & $29.6\%$ & $29.6\%$ \\
At least one distractor plausible & $61.8\%$ & $58.2\%$ \\
Duplicate or overlapping options & $25.5\%$ & $27.3\%$ \\
\bottomrule
\end{tabular}

\end{table}
}

\subsection{Human Evaluation for DG}
Due to the nature of distractor generation, where we do not have access to golden data with "correct" reference outputs (indeed, the goal is to be wrong), we use human raters to evaluate the quality of our model.
Our rater template measures distractor quality along two dimensions:
(1) whether distractors seem plausibly correct to raters (i.e., whether they succeed at "fooling" the rater), and
(2) whether distractors are distinct from each other and from the key.

To measure this, we present the raters with the multiple choice quiz question, without telling them the correct answer. We then ask them to perform the following two tasks:

\begin{enumerate}
   \item Choose the answer you think is correct. If you are unsure, mark all the answer options you think might be correct. (Do not try to research the correct answer.)
   \item Mark all duplicate or overlapping answer options.
\end{enumerate}

\noindent 
To evaluate the effect of fine-tuning on our own question-answer pairs, we ran the same human evaluation on distractors generated by a closed-book question answering model released with~\cite{roberts2020much}. In particular, we used the T5-11B model fine-tuned on TriviaQA as the baseline model. We ran both distractor generation models on the same question-answer pairs generated by our QAG model, and use the same deduplication heuristic (described in Section~\ref{dgdedup}) for the resulting distractor candidates. We note that we could not directly compare our model to previous distractor generation models in the literature because of the uniqueness of our question-answer format which requires potentially long, unstructured answers to free-form questions.

We used 5 raters each to rate 669 sets of answer options, for a total of 3358 ratings. We compute following statistics from this human evaluation:

\begin{itemize}
    \item \textbf{Plausible options per question.} By "plausible," we mean an answer options marked by a rater as potentially correct. Since plausibility is inherently subjective, we consider each rater's ratings independently; the reported numbers are averaged across raters. Since raters are instructed to mark all answer options they think might be correct if they are unsure about the correct answer, the number of plausible answer options per question measures raters' uncertainty in the correct answer.
    \item \textbf{Single plausible option.} The percentage of question-rater pairs where the rater only marked one answer option as correct (i.e., they were fairly certain that that one was the correct answer option).
\end{itemize}

\noindent We use the following four different statistics for measuring how well distractors fooled the raters (for all four, larger numbers are better).

\begin{itemize}
    \item \textbf{Single plausible is a distractor.} Among ratings where the rater was certain about the correct answer, the percentage of times they chose a distractor.
    \item \textbf{Keys not plausible.} The percentage of ratings where the rater did not choose the key as a potentially correct answer option, regardless of whether they marked one or more options as potentially correct. 
    \item \textbf{Distractors plausible.} The percentage of distractor-rating pairs where a distractor was marked as a potentially correct answer option; in other words, the average "plausibility" of distractors.
    \item \textbf{At least one distractor plausible.} The percentage of question-rater pairs where the rater marked at least one of the distractors as potentially correct; in other words, where the rater wasn't certain that the key is the correct answer. This is the most liberal measure of success among the four; it is by definition at least as large as the other three.
\end{itemize}

Finally, to measure duplicate or overlapping options we compute:

\begin{itemize}
    \item \textbf{Duplicate or overlapping options.} As these are objective properties of the answer options, we take the majority vote among the five raters for each answer option and present the percentage of all answer options.
\end{itemize}

\section{Experiment Results} \label{results}

In this section we report the results of both QAG and DG experiments against baselines and ablation studies.

\subsection{Question-Answer Generation}

Table~\ref{tab:qag} shows the performance of various QAG methods on the \emph{NewsQuizQA} test set. Our complete QAG model outperforms all of the baselines evaluated. More specifically, methods using minimum reference loss outperformed methods that attempted to transform the multiple-reference dataset into a single-reference dataset. This improved training objective allowed PEGASUS to outperform T5-3B, a model with approximately 5x more parameters. Mid-training via weak supervision also provided an additional gain. We note the importance of the length normalization factor in encouraging the model to learn longer sequences.

Surprisingly, using PEGASUS, randomly sampling a reference output outperforms the disaggregate-and-shuffle method proposed by previous machine translation works. We hypothesize that this might be due to the nature of QAG, where target outputs are significantly more diverse, possibly interfering with the training dynamics. The improved sample efficiency of recent sequence-to-sequence models may also contribute to this result. Between PEGASUS and T5, T5 outperform PEGASUS in the disaggregated case. However, PEGASUS does competitively when the single reference is randomly sampled. Finally, we note that as intended, there is a significant difference between ROUGE and ROUGE-QAG scores.

\subsection{Distractor Generation}

We report the results of the human evaluations for both our proposed model and the baseline from~\cite{roberts2020much} in Table~\ref{tab:dgeval}.

These human evaluations show that our distractor generation model succeeds at generating high quality distractors.
For a significant majority of questions, human raters thought at least one of the distractors might be a correct answer.
Even when raters were fairly certain in the answer option they chose, almost half of the time they chose a distractor instead of the correct answer.

The results further show that fine-tuning on our own automatically generated question-answer data improves model performance: our model performed at least as well as the model fine-tuned on TriviaQA among all quality dimensions, even though the latter model is an order of magnitude larger in terms of the number of parameters.

\section{Case Study}\label{case_study}

Ultimately, the quality of automatically generated multiple-choice questions must be evaluated holistically on real users. This is a very challenging evaluation task due to the subjective nature of user enjoyment and skill level. As such, we formulate our evaluation as a case study of our automatically-generated News Quiz questions in comparison to a baseline approach for measuring news informedness: a self-assessment quiz, which we treat as control. 

To conduct the case study, we ran weekly News Quiz surveys for nine consecutive weeks on the Google Surveys platform using quiz questions generated by our models. The first quiz was ran on August 11, 2020 and the last quiz was ran on October 7, 2020. At the beginning of each week, we ran our models to generate questions based on prominent news stories about COVID-19 from the past week. We restrict ourselves only to COVID-19 in order to control as many variables as possible, especially as COVID-19 remained the major news story during the duration of our study. 

The questions passed through light curation and editing by a journalism expert before being shown to users. This was an important step for two reasons. First, this guaranteed that no misleading information could be relayed to survey respondents. Second, since we intentionally omit question ranking from the scope of our current work, we are able to rely on curation for support. In order to limit the amount of editing a human curator was allowed to perform: each week the curator received ten multiple-choice questions with five distractor candidates each. The curator is allowed to pick three questions out of the ten, and three distractors out of five distractor candidates for each question. The curator is then restricted to only to three categories of edits: (1) minor fixes to grammar and spelling, (2) insertion of extra information to clarify news sources and dates, and (3) modification of distractors to have consistent formatting, e.g. if numbers represent percentages all distractors should have percentages symbols. Throughout the entire study, only 3 question and answers out of 27 selected, and 26 distractors out of 81 selected were edited. Figure~\ref{fig:example} shows an example of an unedited multiple-choice question generated by our system.

From these three curated questions each week, we construct a News Quiz suitable for the Google Surveys platform. This News Quiz consisted of a screening question, instructions for the quiz with an option to opt-out, the three curated multiple-choice questions, followed by the correct answers for the questions and a question asking respondents to report how many of them were answered correctly. Finally, to gauge users' satisfaction with our quizzes and their potential effect on users' news informedness, we ask the following three questions (each on a five-point scale from "Not at all" to "Very much so"):

\begin{itemize}
   \item Q1: To what extent did this quiz help you learn more about this news topic (COVID-19)?
   \item Q2: To what extent did this quiz make you want to read more about this news topic (COVID-19)?
   \item Q3: Would you want this type of quiz to be a part of your regular news reading experience?
\end{itemize}

\begin{figure}[t!]
  \centering
    \includegraphics[width=\linewidth]{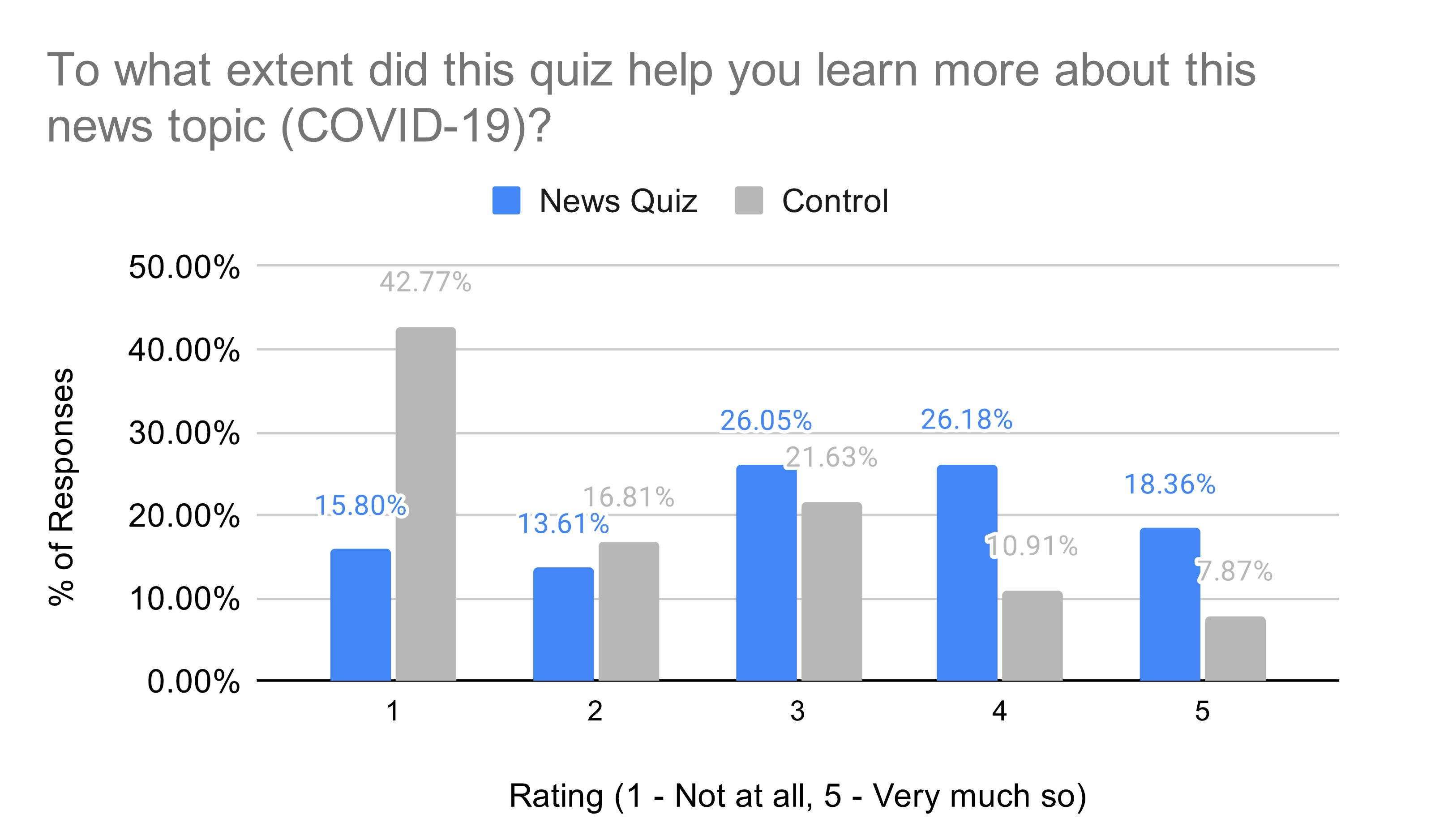}
    \includegraphics[width=\linewidth]{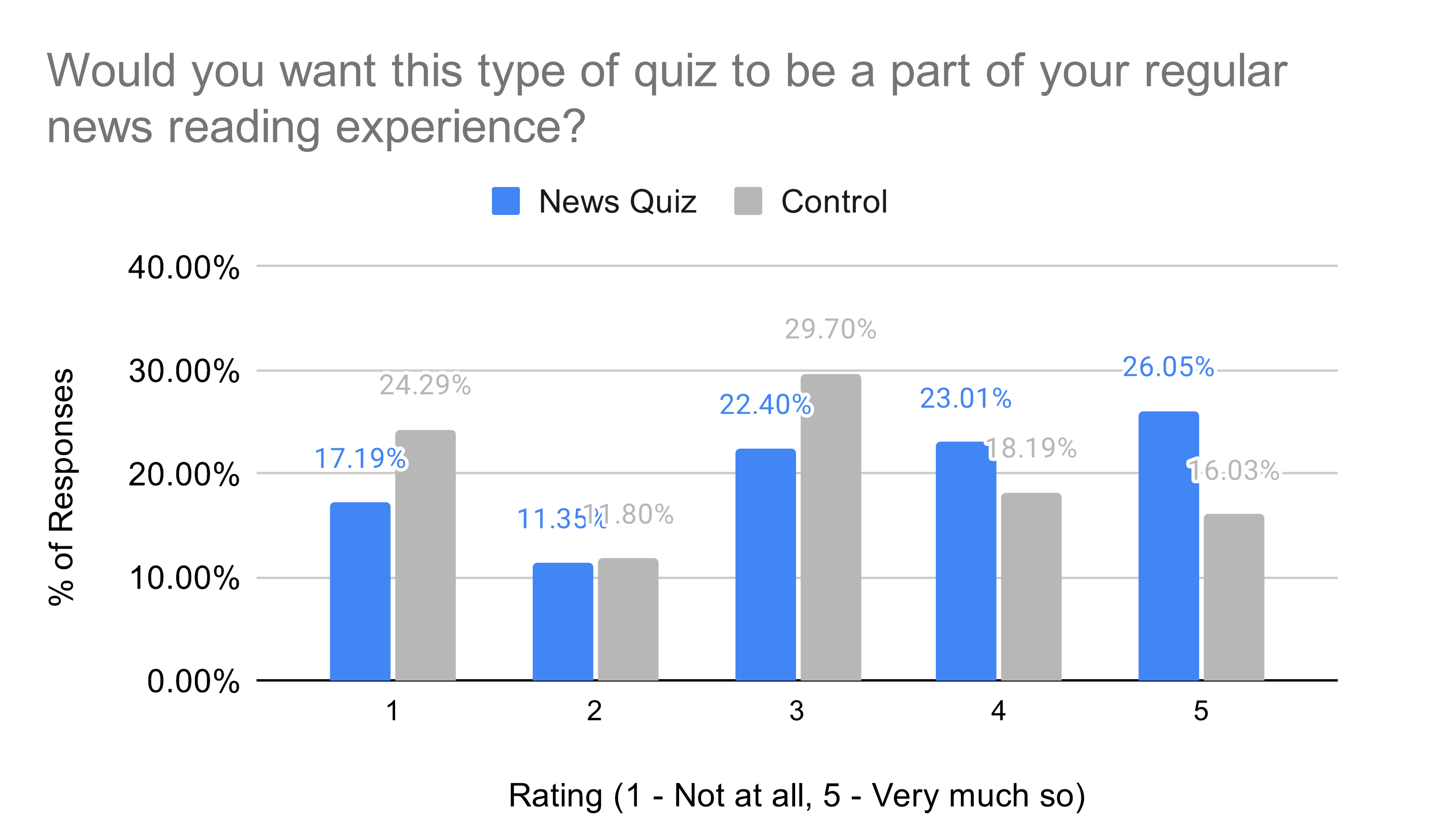}
  \caption{Comparison of responses for two selected questions asked in the Google Surveys case study.}
  \Description{Bar charts showing the distribution of ratings for the questions, "To what extent did this quiz help you learn more about this news topic (COVID-19)?" and "Would you want this type of quiz to be a part of your regular news reading experience?", for both the news quiz and the control survey.}
  \label{fig:survey}
\end{figure}

\noindent For the control, we maintain exactly the same structure, except the three multiple-choice questions were replaced with questions that asked the user to self-assess their news informedness. In particular: (1) "How often do you read news by searching the topic on search engines?" (2) "How often do you read news by using a news aggregator site or app?" (3) "How often do you read news by going directly to a specific publisher?" and (4) "How informed are you about news stories related to COVID-19?"

Since the control quiz does not rely on evolving news information, the control was ran only once with 1,000 responses collected. For the News Quiz, 1,000 responses were collected each week for nine consecutive weeks. Respondents were restricted to the “Android-smartphone users" option in Google Surveys, which targets users of the Google Opinion Rewards app.

The result of this experiment showed News Quiz outperforming the control on all three rating questions: $3.18 \pm 0.03$ vs. $2.24 \pm 0.08$ for Q1, $3.02 \pm 0.03$ vs. $2.47 \pm 0.08$ for Q2, and $3.29 \pm 0.03$ vs. $2.90 \pm 0.08$ for Q3. Although, this result was somewhat expected as the control does not directly contain any information about the news topic, it does suggest that our automatically generated multiple-choice questions are effective. 44\% of users felt positively (4- and 5-ratings) that the News Quiz helped them learn more about the news topic, while 49\% of users, with a plurality of 5-ratings, felt positively about incorporating such a quiz as part of their regular news reading experience (Figure~\ref{fig:survey}). Given the subjective nature of reading preferences, knowledge level, and user enjoyment, we consider this a positive result.

Users on average answered 1.66 out of 3 questions correctly, and the weekly average did not fluctuate more than 0.6 from this global average weekly. This result suggests that our process on consistently produced medium difficulty questions. Given this, our method may be effective in future work to measure news informedness between different populations in an A/B test setting.

\subsection{Example Quiz}
Here we include an example three-question quiz from our study, generated by our system and curation process on September 9, 2020. Insertions made by the curator are marked in green, while deletions are marked in red and are crossed out. All other text was generated by our models as-is. We also include for each example the generated summary that served as input to the question-answer generation model. Respondents did not have access to this summary text.

\begin{figure}[h!]
  \centering
    \includegraphics[width=\linewidth]{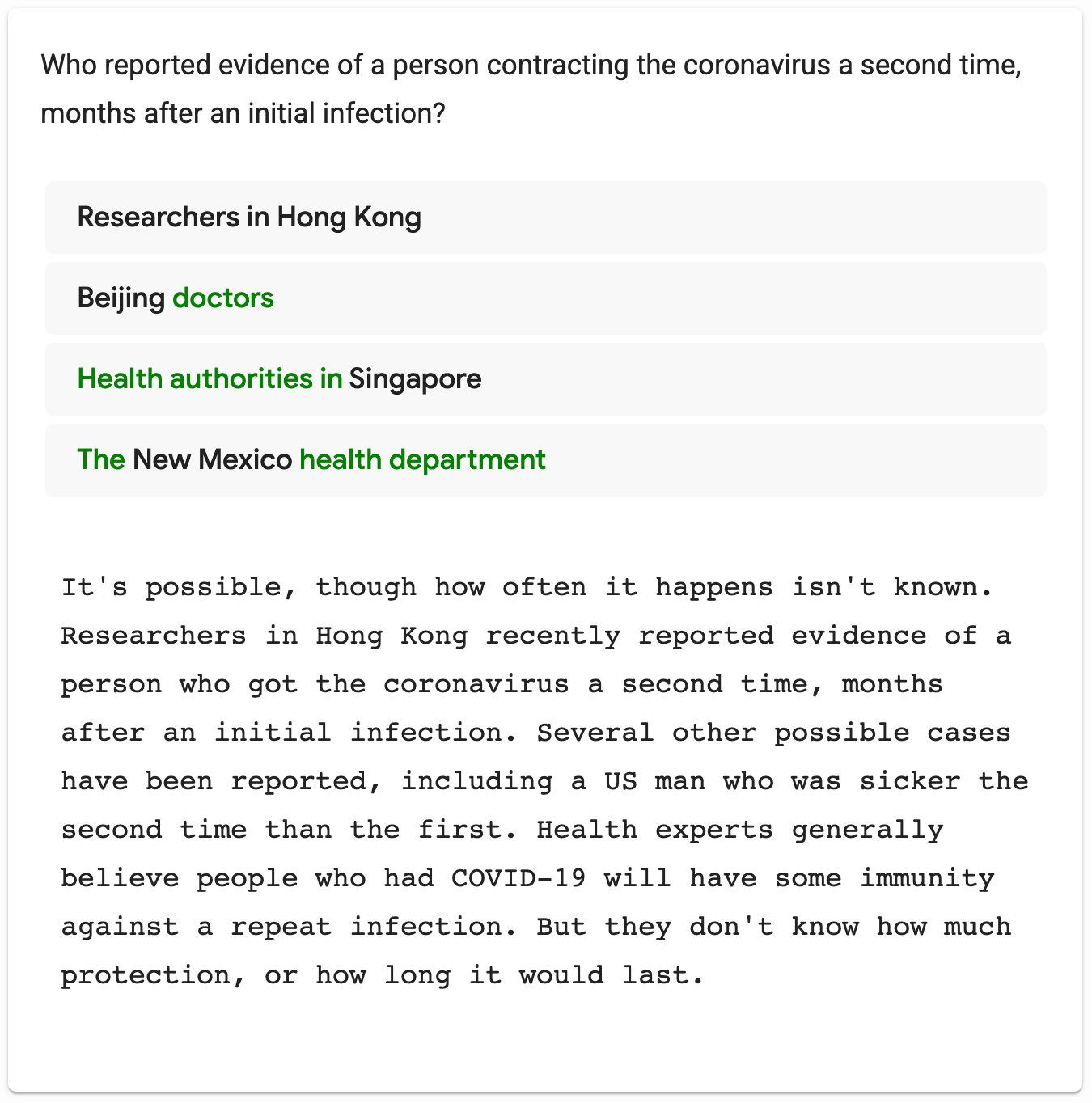}
  \caption{Question 1, the answer is "Researchers in Hong Kong".}
  \Description{Question: Who reported evidence of a person contracting the coronavirus a second time, months after an initial an initial infection? Answer option 1: Researchers in Hong Kong. Answer option 2: Beijing doctors. Answer option 3: The New Mexico health department. Answer option 4: Health authorities in Singapore.}
  \label{fig:quizexample1}
\end{figure}

\begin{figure}[h!]
  \centering
    \includegraphics[width=\linewidth]{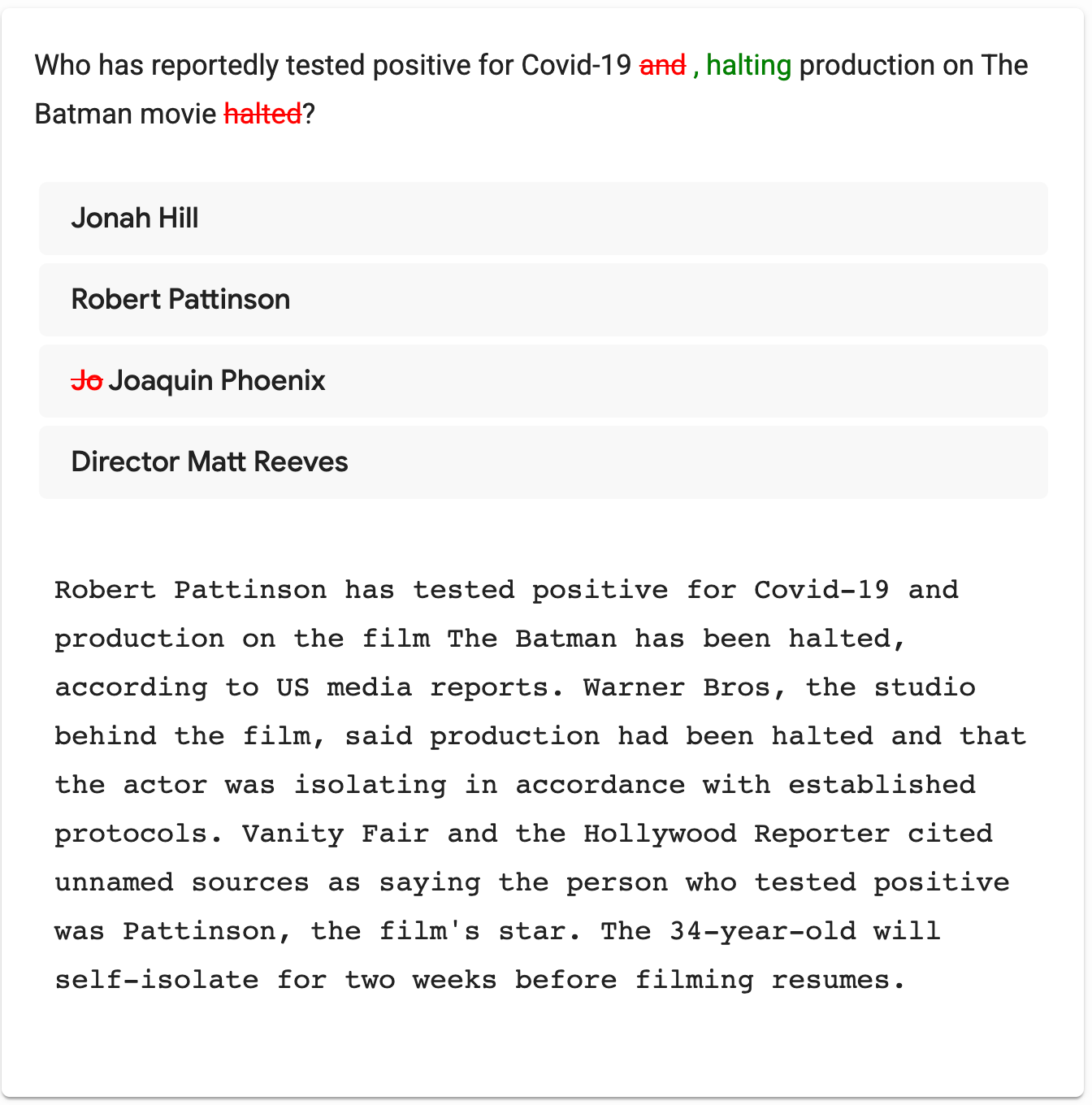}
  \caption{Question 2, the answer is "Robert Pattinson".}
  \Description{Question: Who has reportedly tested positive for Covid-19, halting production on The Batman movie? Answer option 1: Jonah Hill. Answer option 2: Robert Pattinson. Answer option 3: Joaquin Phoenix. Answer option 4: Director Matt Reeves.}
  \label{fig:quizexample2}
\end{figure}

\begin{figure}[h!]
\bigskip
\bigskip
\bigskip
  \centering
    \includegraphics[width=\linewidth]{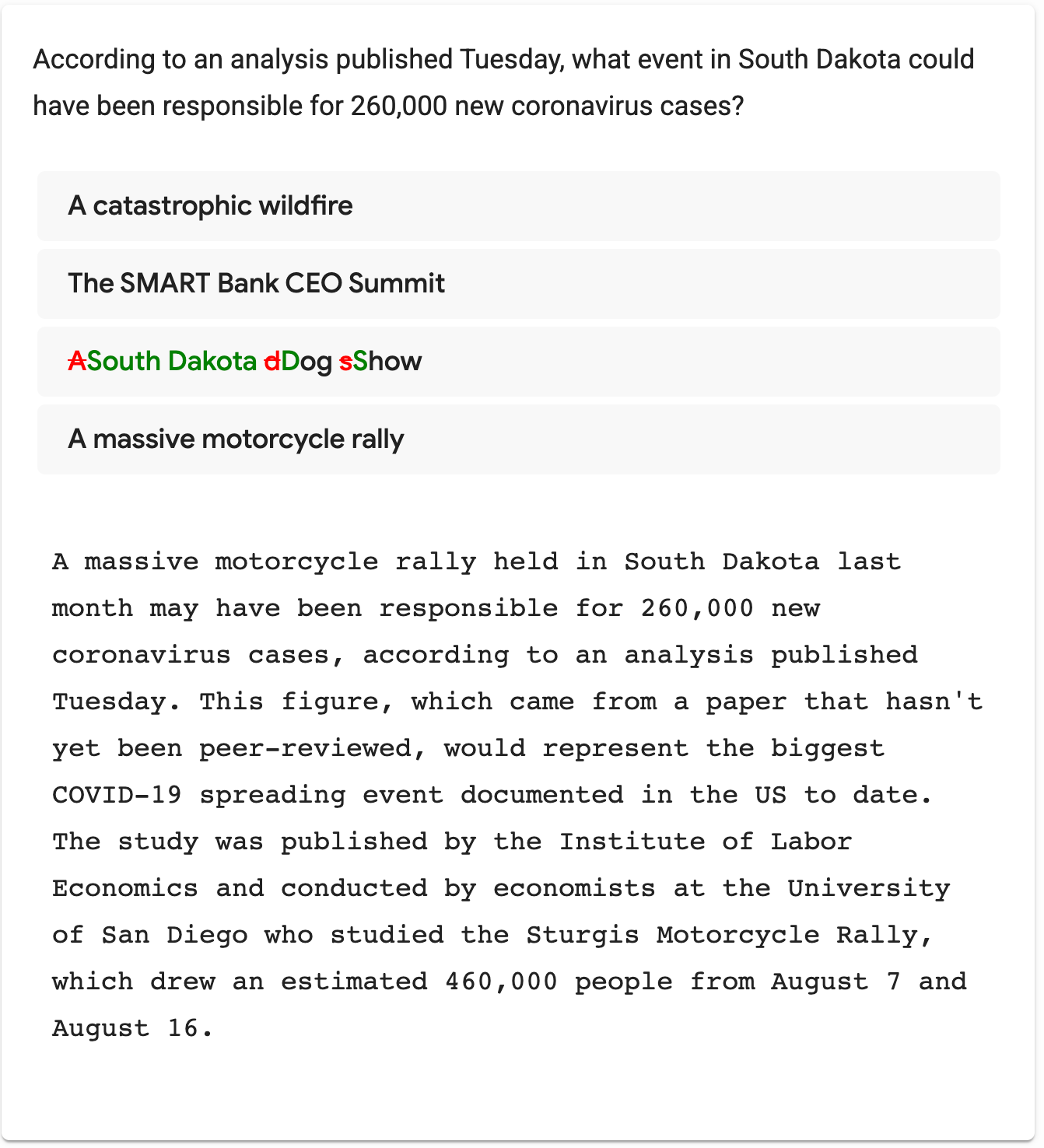}
  \caption{Question 3, the answer is "A massive motorcycle rally".}
  \Description{Question: According to an analysis published Tuesday, what event in South Dakota could have been responsible for 260,000 new coronavirus cases? Answer option 1: A catastrophic wildfire. Answer option 2: The SMART Bank CEO Summit. Answer option 3: South Dakota Dog Show. Answer option 4: A massive motorcycle rally.}
  \label{fig:quizexample3}
\end{figure}

Although all three questions for this week's quiz required some editing, all edits were relatively minor, often stylistic. In Question 1 (Figure \ref{fig:quizexample1}), distractors were adjusted to have consistent specificity. This is a challenging problem;  our current method generates distractors independently of each other. In Question 2 (Figure \ref{fig:quizexample2}), the grammar of the question was altered to sound more natural and an extraneous token was removed from the third distractor. Finally in Question 3 (Figure \ref{fig:quizexample3}), a distractor was adjusted to appear more specific. This was also a stylistic change similar to the one in Question 1. We leave improvements in distractor consistency as an area for future work. 

\section{Conclusion}
In this work, we propose automatically generating multiple choice quiz questions to measure users' informedness about the news.
We introduce novel approaches for training natural language generation models to solve this problem. We also introduce a new dataset, \emph{NewsQuizQA}, consisting of human-written question-answer pairs about news article summaries, which we used for training and evaluating our models.
We demonstrate the performance of our models using both automated metrics and human evaluations, as well as using a case study with real users on the Google Surveys platform.

\begin{acks}
We would like to thank Kerri Connolly for her journalistic guidance for the project, Catherine Cloutier and Karina Martinez-Carter for organizing the review of our dataset, and Colin Keogh, James Reilly, and William Vambenepe for their valuable comments and suggestions.
\end{acks}

\bibliographystyle{ACM-Reference-Format}
\bibliography{main}
\end{document}